\documentclass{article}

     \usepackage[preprint]{nips_2018}

\usepackage[utf8]{inputenc} 
\usepackage[T1]{fontenc}    
\usepackage{hyperref}       
\usepackage{url}            
\usepackage{booktabs}       
\usepackage{amsfonts}       
\usepackage{nicefrac}       
\usepackage{microtype}      
\usepackage{amsfonts,amsmath,amssymb,array}
\usepackage{makeidx}         
\usepackage{graphicx}        
\usepackage{multicol}        
\usepackage[bottom]{footmisc}

\usepackage{times}
\usepackage{parskip}

\usepackage{wrapfig}
\usepackage{subcaption}

\usepackage{amssymb}
\usepackage{epsfig}
\usepackage{amsfonts,amsmath,amssymb,array}
\usepackage{tabularx}   
\usepackage{colortbl}
\usepackage{longtable}
\usepackage{booktabs}
\usepackage{url}
\usepackage{icomma}
\usepackage{pstricks}
\usepackage{longtable}
\usepackage{multirow}
\usepackage{color}
\usepackage{algorithm2e}

\usepackage{subfiles}
\usepackage{float}
\usepackage{mathtools}

\title{Neural Logic Rule Layers}

\author{%
  Jan Niclas Reimann\\
  Department of Automation Technology\\
  South Westfalia University of Applied Science\\
  Soest, Germany \\
  \texttt{reimann.janniclas@fh-swf.de} \\
 \And
  Andreas Schwung \\
	Department of Automation Technology\\
  South Westfalia University of Applied Science\\
  Soest, Germany \\
  \texttt{schwung.andreas@fh-swf.de} \\
}

\begin{document}

\maketitle

\begin{abstract}
Despite their great success in recent years, deep neural networks (DNN) are mainly black boxes where the results obtained by running through the network are difficult to understand and interpret. Compared to e.g. decision trees or bayesian classifiers, DNN suffer from bad interpretability where we understand by interpretability, that a human can easily derive the relations modeled by the network. A reasonable way to provide interpretability for humans are logical rules. In this paper we propose neural logic rule layers (NLRL) which are able to represent arbitrary logic rules in terms of their conjunctive and disjunctive normal forms. Using various NLRL within one layer and correspondingly stacking various layers, we are able to represent arbitrary complex rules by the resulting neural network architecture. The NLRL are end-to-end trainable allowing to learn logic rules directly from available data sets. Experiments show that NLRL-enhanced neural
networks can learn to model arbitrary complex logic and perform arithmetic operation over the input values.  
\end{abstract}

\section{Introduction}

Deep neural networks (DNN) have shown great success in previous years in various domains like image classification, speech recognition and natural language processing. However, DNNs operate in a black box fashion where the results obtained by running through the network are in general difficult to understand. Hence, compared to other machine learning algorithms like decision trees or fuzzy classifiers, DNN suffer from low interpretability. However, for various applications, a reasoning structure which can easily be understood by a human is highly desirable. A natural form to constitute such easily understandable relations are provided by applying logical rules. 

In this paper, we develop new learning modules for neural networks called neural logic rule layers (NLRL) with in-built capacity to represent logical rules. The approach relies on the concept of boolean logic which represents rules in form of algebraic equations. As every boolean logic can be represented by a combination of negations, AND and OR-rules, we concentrate on a neural network representation of these logic subtypes, i.e. we propose neural network layers embedding disjunctive and conjunctive normal forms if at least two layer are stacked. The NLRL is split into three blocks, namely the negation operator realized using a gating mechanism, the AND-connector and the OR-connector. The AND- and OR-connector thereby share their weights. Optionally, the decision whether an AND- or an OR-clause is used for reasoning can be represented by an additional gating unit. Despite the combinatorial character of the logic units, the proposed architecture is end-to-end trainable using backpropagation.

Besides their increased interpretability, NLRL inherently incorporate inductive bias into the neural network structure. Due to the restrictions raised on the network parameters for representing logical rules, comprehensible inductive bias is introduced, especially if used as a classification layer. Furthermore, if knowledge is available about the underlying classification problem, NLRL allow for the direct incorporation of such knowledge in form of logical rules. The rules could finally be fine tuned using backpropagation. Furthermore, if the NLRL is used in the output layer, i.e. the output is the result of a single rule evaluation, it is most likely that no output neuron fires if an unknown input is presented. This is in sharp contrast to the most often used softmax layer where the most likely class is output yielding misleading results.  

We experiment on a variety of different logic function and different network structures. Furthermore, we test on various nonlinear function representations which are also representable by the NLRL. The results obtained on the synthetic data sets are very encouraging motivating further investigations on e.g. image classification.

\section{Neural Logic Normal Form Units}

We propose neural logic normal form units, i.e. units to represent logical relations in form of AND- and OR-rules. We model the relations by using the framework of propositional logic and boolean algebra as discussed next. 

\subsection{Propositional Logic and boolean algebra}

Logical rules provide a convenient way to model expert knowledge in various domain. Such rules can be described using basic operations as logical AND (conjunction), logical OR (disjunction) and negation using the truth tables TRUE and FALSE. Equivalently, we can represent the truth values by integer values 0 and 1 and apply corresponding arithmetic operation. Hence, assuming integer variables $x,y \in \{0,1\}$, the conjunction, disjunction and negation can equivalently be defined by the algebraic relations  
\begin{align}
x\wedge y&=xy,\\
x\vee y&=x+y-xy,\\
\neg x&=1-x.
\end{align}
The generalization of the conjunction and disjunction to more than two input dimensions, i.e. $x \in \{0,1\}^n$, yields
\begin{align}
\bigwedge^n_{i=1} x_i &= \prod^n_{i=1} x_i,\\
\bigvee^n_{i=1} x_i &= \left(\bigodot^n_{i=2} (1 \ \  -x_i) \odot (-1 \ \  x_1)\right)\mathbf{1}+1. 
\end{align}
where the Kronecker operator is used to represent the algebraic OR. Based on the basic operations, additional operations can be defined where implication, exclusive OR and equivalence being the most prominent ones defined as follows
\begin{align}
	x\rightarrow y &=\neg {x}\vee y \\
	x\oplus y &= (x\vee y)\wedge \neg {(x\wedge y)} = (x\wedge \neg y)\vee (\neg x\wedge y)\\
	x\equiv y &= \neg {(x\oplus y)} = (x\wedge y)\vee (\neg x\wedge \neg y)
\end{align}
As indicated, the above additional operations can be traced back to the basic operations. However, they require a sequence of conjunction and disjunction operations or vice versa. 

Furthermore, conjunction, disjunction and negation are related via the De-Morgan laws
\begin{align} 
\bigwedge^n_{i=1} x_i &= \neg (\bigvee^n_{i=1} \neg x_i),\\
\bigvee^n_{i=1} x_i &= \neg (\bigwedge^n_{i=1} \neg x_i).
\end{align}
As a consequence, for arbitrary logical relations, only two of the three basic operations are required. This property will be further exploited as an alternative representation for the logical layer definition discussed next.

\subsection{End-to-end training of logic layers}

\begin{figure}[h]
 \centering
 \includegraphics[width=\columnwidth,keepaspectratio]{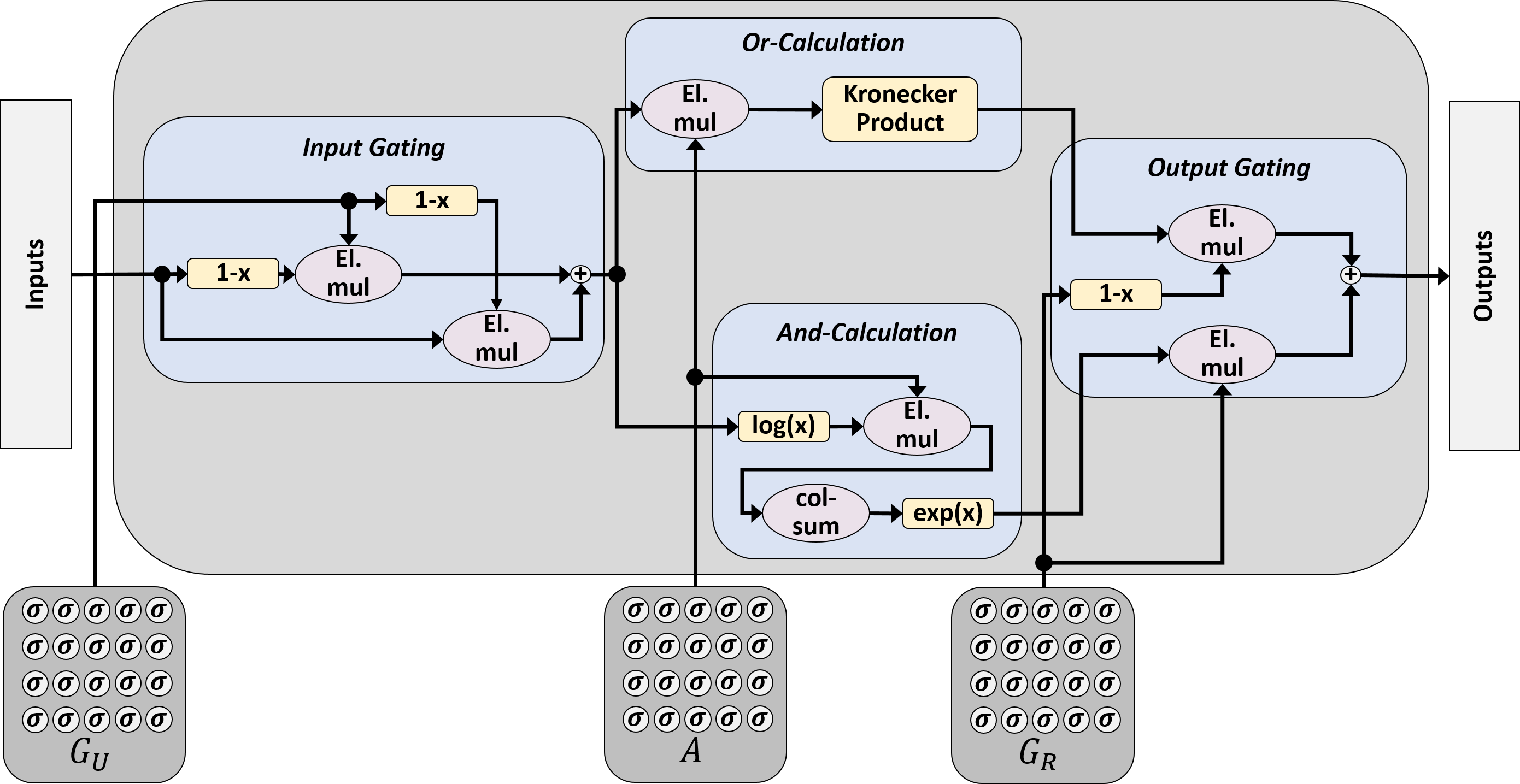}
\caption{Architecture of NLRLs.}
\label{fig:NLNFUarchi}
\end{figure}

Using boolean algebra, rules can be equivalently defined by means of algebraic equations. To represent boolean algebra in form of a neural network architecture, we propose a corresponding layer structure consisting of three blocks, namely the input negation, a parallel path to represent the AND and OR-relation which share their weights and an output gating to decide whether AND or OR-relation shall be used. The architecture is illustrated in Fig.~\ref{fig:NLNFUarchi}.

For the feedforward path we obtain the following description:
\begin{align}
	\text{Negation Gating:} & \quad \hat{x}=(1-\sigma(g^u)) \circ x+\sigma(g^u) \circ (1-x),\\
	\text{AND-rule:} & \quad \text{AND}= \text{exp} (\mathbf{A}(\text{log}(|\hat{x}|+\epsilon))),\\
	\text{OR-rule:} & \quad \text{OR}= ((1 \ \  -a_n\hat{x}_n)\odot \ldots \odot (1 \ \  -a_2\hat{x}_2) \odot (-1 \ \  a_1\hat{x}_1))\mathbf{1}+1,\\
	\text{Output Gating:} & \quad y=(1-\sigma(g^r)) \circ \text{AND}+ \sigma(g^r) \circ \text{OR}.
\end{align}
where $\mathbf{A}$, $\mathbf{g}^u$ and $\mathbf{g}^r$ are the learnable weights. The weight vector $\mathbf{g}^i$ models the negation of the logical input variable while $\mathbf{g}^o$ is the output gating, which decides whether conjunction or disjunction are used by the neurons of the layer. The component $a_{i,j}$ of the weight matrix $\mathbf{A}$ assigns the input variable $x_i$ to the $j$th rule in the layer. Note that all weights have to be restricted to $[0,1]$. To allow for an end-to-end training of the variables, we define the weights as continuous variables which are send subsequently through a sigmoidal activation following the same approach as in~\citet{Trasky2018}. Note also, that the logic relations as modeled by the forward path hold only if the inputs to the layer are boolean values, i.e. $x \in \{0,1\}^n$. However, similar to the weights, to assure an end-to-end training we use sigmoidal activations in the preceding layer to restrict the values to the interval $[0 \ 1]$. 

As previously discussed, we can restrict ourselves to the negation normal form~\citep{Robinson2001}, i.e. the use of negation operator together with either conjunction or disjunction operator to represent arbitrary logical rules. Hence, using the same architecture as in Fig.~\ref{fig:NLNFUarchi}, we can alternatively implement either the conjunction or disjunction path alone. In this case, the output gating is replaced by a negation gating of the output variable. This latter representation provides some computational advantages, as the parallel computation of conjunction and disjunction is avoided. Particularly, due to the Kronecker product, the calculation of the disjunction scales exponentially with the input dimension and the number of neurons in the layer and is hence computationally expensive. Hence, we always use the conjunction operator in this setting.

\subsection{Representational Power of Stacked NLL}

We already discussed the representation of a single NLRL-neuron as either representing a conjunction or disjunction rule which is made learnable by the output gating. Alternatively, we can implement the negation normal form by using the conjunction solely and allow for negation of the output. Also, we can fix the representation by design, i.e. omit the output gating completely. If NLRL are stacked layer-wise, the representational power drastically increases. Stacking two layers of NLRLs allows for arbitrary logic relations in form of logic normal forms, i.e. either conjunctive or disjunctive normal forms which particularly includes the previously defined implication, exclusive or and equivalence operators. 
 
As a by-product, the NLRLs are also able to represent various forms of functions. Due to the multiplicative nature of the algebraic AND-rule, multiplication of inputs as well as polynomial relations, where the maximum achievable order of the polynomial equals the number of layers, are learnable. The OR-rule can also represent addition and substraction operations. Combining both algebraic AND and OR-rule, arbitrary complex relations combining the above functions can be learned.

The constraints on the optimization parameters and the in-built rule base structure of NLRL already provide considerable inductive bias to the learning process when NLRL are used. However, there are application examples where additional a priori knowledge is available, e.g. by experts or experienced users. Sharing this additional knowledge with the learning process can be beneficial. Human knowledge can often be represented by logic rules or fuzzy rules, hence NLRL naturally allow for a direct incorporation of such rules by correspondingly initializing the optimization parameters instead of random initialization.

Furthermore, as we use differentiable activation functions for the inputs, they are restricted to the interval $[0\ 1]$. By interpreting the input activations as membership functions, the corresponding NLRL can be interpreted as representing a fuzzy rule. 

Finally, we consider the extrapolation behavior of NLRLs. To this end we consider a typical classification scenario where we assume that a feature extractor, e.g. some convolutional layers, have been stacked downstream of the logic layers. We compare the behavior of NLRL with classical fully connected layers with a final softmax layer for multiclass classification. If we assume that an input is provided to the network, that does not belong to any of the previously seen classes, e.g. an outlier, the softmax layer will output the most likely, but wrong class affiliation. Hence, the result could be misleading. In contrast, the NLRL will simple give no output, i.e. no rule will be fired, such that the results can be directly interpreted as a not classified input.

\section{Related Work}

Our rule base architecture in NN can be classified to the research direction of informed machine learning, see~\citet{Rueden2019} for a recent survey. Informed machine learning summarizes approaches of incorporating various forms of a priori available domain knowledge into the learning process which constitutes a form of inductive bias~\citet{Desjardins1995}. The approaches differ in the type of knowledge (e.g. natural sciences, world knowledge or expert intuition), the way the knowledge is represented (e.g. rules, constraints, simulations or graphs) and where the knowledge is incorporated into the learning pipeline (e.g. training data, training algorithm or hypothesis space)~\citep{Rueden2019}. Our NLRL incorporate world and expert knowledge in form of rules into the hypothesis space. Related approaches also integrating logical rules have already been presented. In~\citet{Hu2016}, a regularization of NN is performed by using logical relations as regularizer while~\citet{Diligenti2017} incorporate logic rules in form of constraints into the training scheme. Similarly, \citet{Diligenti2017b} propose semantic based regularization. First approaches working directly in the hypothesis space are Markov Logic Networks~\citep{Richardson2006} where a first order logic base is augmented with trainable weights. A hypothesis space approach for fast construction of data sets for machine learning problems has been presented in~\citet{Ratner2016}, while~\citet{Sachan2018} present a similar approach for solving physics problems in text books. In contrast to the previously mentioned approaches, we propose to integrate rules in the hypothesis space by biasing the NN to represent logical rules. This allows for a direct end-to-end learnable integration of logical rules in NN which to the best of our knowledge has not been presented previously.

Another branch in the literature is the development of networks structures for logic testing with application in software testing~\citep{Evans2018}. In~\citet{Allamanis2017} continuous semantic representations of symbolic expressions are learned using tree based NN which represent logic relations. \citet{Evans2018} investigate the ability of NN to model and understand logic entailments. \citet{Selsam2019} propose to learn solvers for propositional satisfiability using NN. However, both network architecture as well as application domain, i.e. testing logical terms vs. training logical relations, considerably differ from the NLRL. We leave the question whether NLRL can also be applied to these problems for future research.

A different branch of research related are binarized neural networks where weights and activations are binary variables~\cite{Hubara2016} which particularly allows for implementation on resource constraint systems which is their main focus. However, modeling logic relations has not been reported for binary neural networks. The proposed NLRL share some common ideas with neural arithmetic logic units proposed in~\citet{Trasky2018}. Particularly, the set up of end-to-end training is similar for both architectures. However, NALUs are designed for basic arithmetic operation like addition and multiplication while we focus on representation of propositional formulas which are not representable with NALUs. 

Due to the rule structure and the input activations which can be interpreted as membership values, NLRL have some relations to fuzzy systems. Especially, in the field of neuro-fuzzy systems~\citet{Shihabudheen2018}, various architectures combining fuzzy logic and neural networks have been proposed where, however, most of the architectures are based on single layer radial basis function networks. Furthermore, training of these networks is restricted to the parameters of the input activations and the rule weighting while rule generation is done via heuristics~\citet{Mitra2000}. In contrast, we propose an end-to-end trainable network architecture where all components of the rule are trainable which has not been done before in the neuro-fuzzy context.

\section{Experiments}

We present preliminary experiments on a given set of elementary logic and arithmetic operations and leave further experiments on e.g. classification tasks for future work. To this end we consider two input variables $x$ and $y$ and perform the basic logic operations $AND$, $OR$, $NOT$ and $XOR$, as well as multiplication, addition, identity operation and constant functions, see Tab.~\ref{tab:resultsfunctions}. 
\begin{table}[h]
  \centering
  \begin{tabular}{ |c|l|l|l|l|l| }
     \hline
     Function & \begin{tabular}{@{}c@{}}$CS=2$, \\2-2-2-2-10,\\ AND-NONEG\end{tabular} & \begin{tabular}{@{}c@{}}$CS=4$,\\ 2-4-4-10,\\ AND-NONEG\end{tabular} & \begin{tabular}{@{}c@{}}$CS=6$,\\ 2-6-6-10,\\ AND-OR\end{tabular} & \begin{tabular}{@{}c@{}}$CS=8$,\\ 2-8-8-8-10,\\ AND-NONEG\end{tabular} & \begin{tabular}{@{}c@{}}$CS=10$,\\ 2-10-10-10,\\ AND-NEG\end{tabular} \\
     \hline
		   $x/2+y/2$              & 71.9\% & 100\% & 100\% & 100\% & 99.7\% \\
       $x\cdot\neg y$     & 99.8\% & 100\% & 93.2\% & 100\% & 100\% \\
       $x\ \text{AND}\ y$ & 93.7\% & 100\% & 99.3\% & 100\% & 99.6\% \\
       $x\ \text{OR}\ y$  & 57.0\% & 100\% & 97.1\% & 99.7\% & 98.4\% \\
       $x\ \text{XOR}\ y$ & 32.6\% & 78.8\% & 88.1\% & 80.3\% & 75.0\% \\
       $x$                & 90.2\% & 100\% & 91.1\% & 99.9\% & 100\% \\
       $y$                & 100\%  & 100\% & 100\% & 100\% & 100\% \\
       $\neg x$           & 94.3\% & 100\% & 99.9\% & 100\% & 100\% \\
       $\neg y$           & 100\%  & 100\% & 100\% & 100\% & 100\% \\
       $0.7$              & 100\%  & 100\% & 100\% & 100\% & 100\% \\
			\hline
			 Overall Acc.   & 83.95\% & 97.88\% & 96.87\% & 97.99\% & 97.27\% \\
     \hline
  \end{tabular}
\caption{Individual scores for the considered logical and arithmetic operations with different network architectures. For each CS, only the best performing network is reported.}
  \label{tab:resultsfunctions}
\end{table}
\begin{table}[h]
  \centering
  \begin{tabular}{ |c|l|l|l| }
     \hline
     Architecture & AND-OR & AND-NEG & AND-NONEG  \\
     \hline
		    2-2-2-10    & 69.54\% & 73.80\% & 83.86\%  \\
       2-2-2-2-10   & 75.59\% & 75.42\% & 83.95\% \\
       2-4-4-10     & 97.77\% & 90.67\% & 97.88\% \\
			2-4-4-4-10    & 87.36\% & 90.19\% & 97.12\% \\
       2-6-6-10     & 96.87\% & 93.92\% & 94.15\% \\
       2-6-6-6-10   & 96.56\% & 90.95\% & 93.87\% \\
			2-8-8-10      & 96.95\% & 96.51\% & 94.40\% \\
       2-8-8-8-10   & 97.05\% & 93.16\% & 97.99\% \\
       2-10-10-10    & 96.91\% & 97.27\% & 94.28\% \\
			2-10-10-10-10  & 95.42\% & 69.23\% & 94.43\% \\
     \hline
  \end{tabular}
\caption{Overall accuracy for the different network architectures, connection sizes and number of layers.}
  \label{tab:resultsfunctions}
\end{table}
The data set is generated by sampling random variables $x$ and $y$ in the interval $[0, \ 1]$ and evaluating the above functions such that 10 output values, one per function, are obtained. We sample a data set of size 100000, where 10000 are hold out for testing.

\begin{figure}
  \begin{subfigure}[t]{.48\textwidth}
    \centering
    \includegraphics[width=\linewidth]{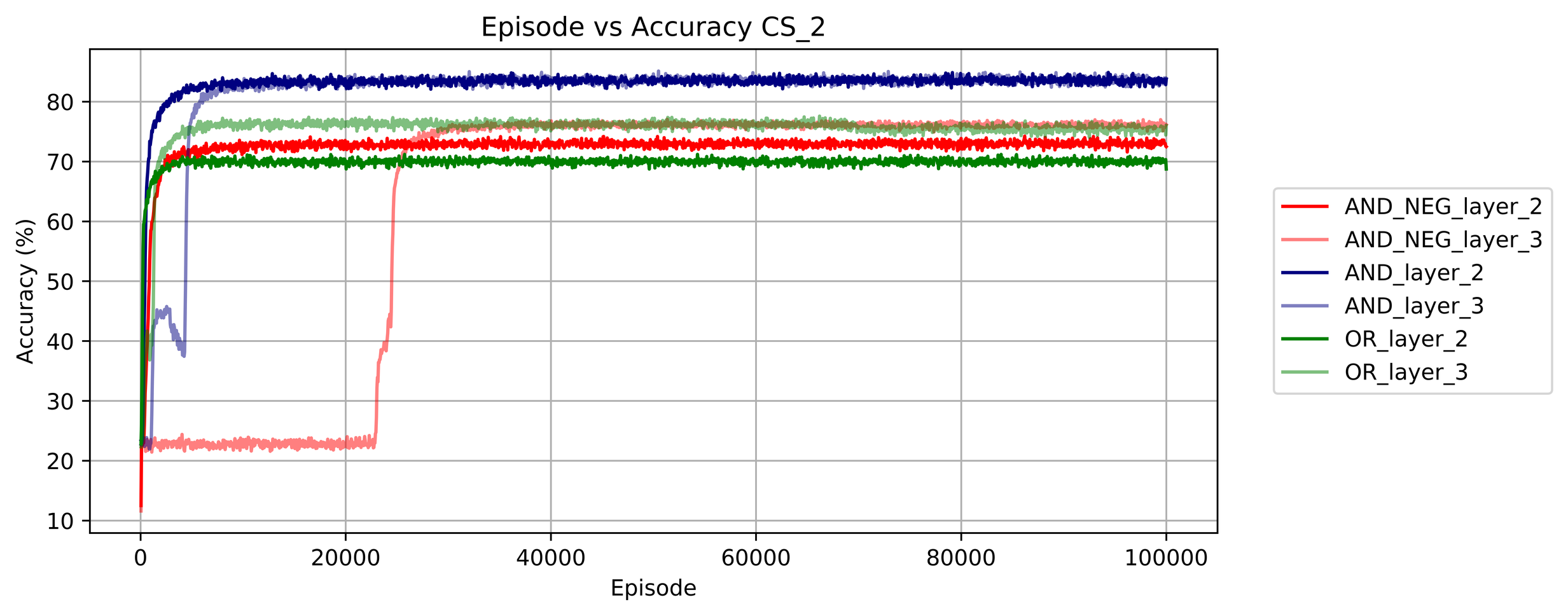}
    \caption{Comparison of testing accuracies for connection size 2.}
  \end{subfigure}
  \hfill
  \begin{subfigure}[t]{.48\textwidth}
    \centering
    \includegraphics[width=\linewidth]{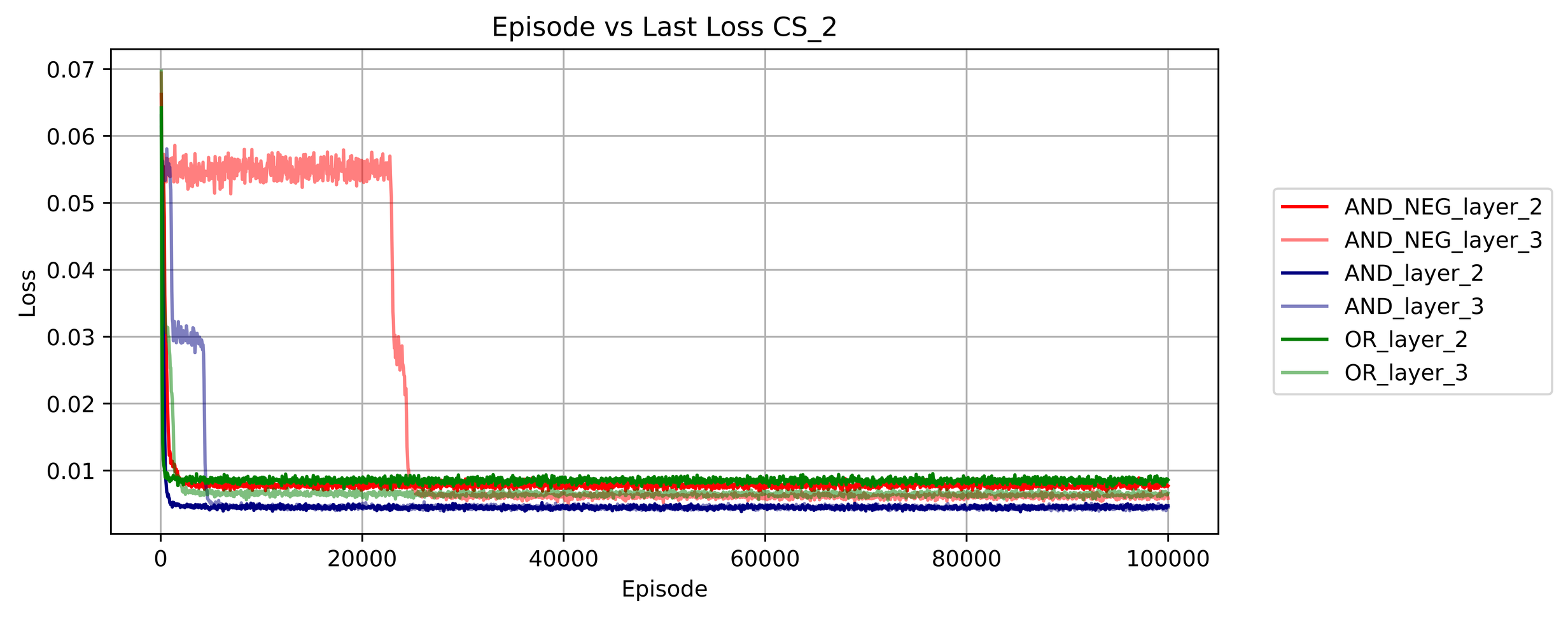}
    \caption{Comparison of loss functions for connection size 2.}
  \end{subfigure}

  \medskip

  \begin{subfigure}[t]{.48\textwidth}
    \centering
    \includegraphics[width=\linewidth]{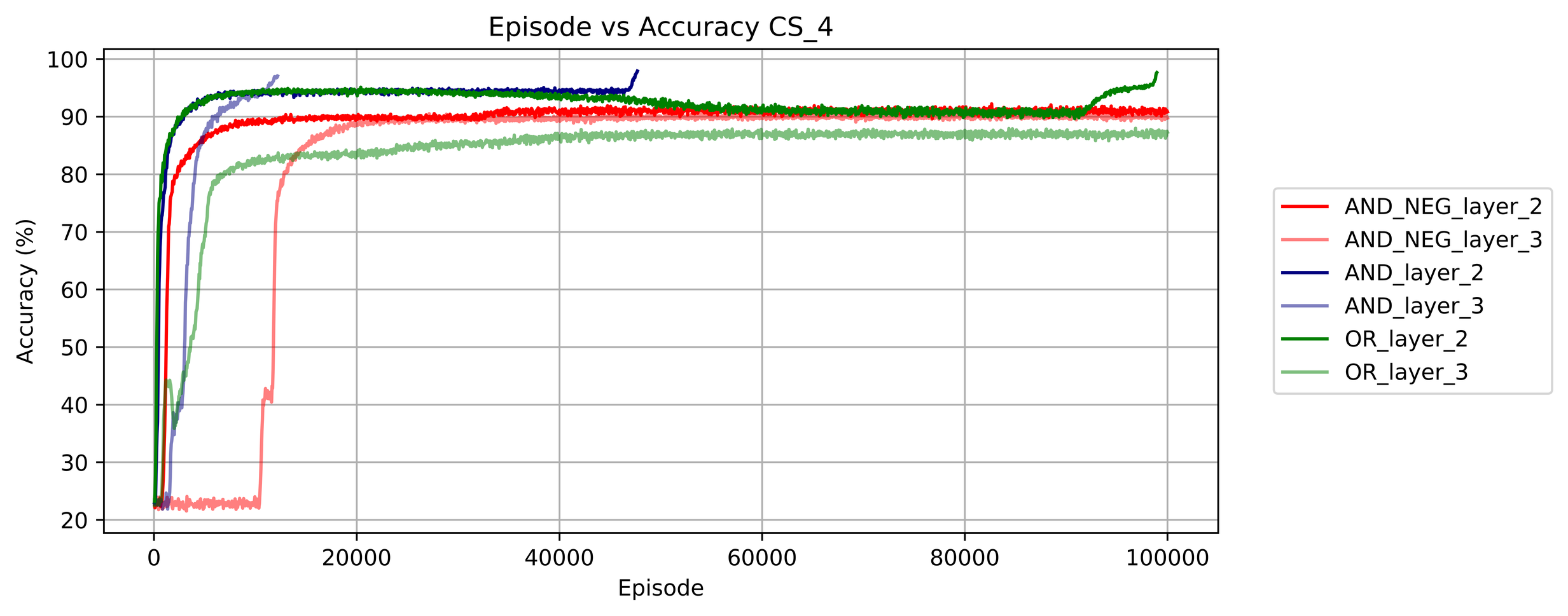}
    \caption{Comparison of testing accuracies for connection size 4.}
  \end{subfigure}
  \hfill
  \begin{subfigure}[t]{.48\textwidth}
    \centering
    \includegraphics[width=\linewidth]{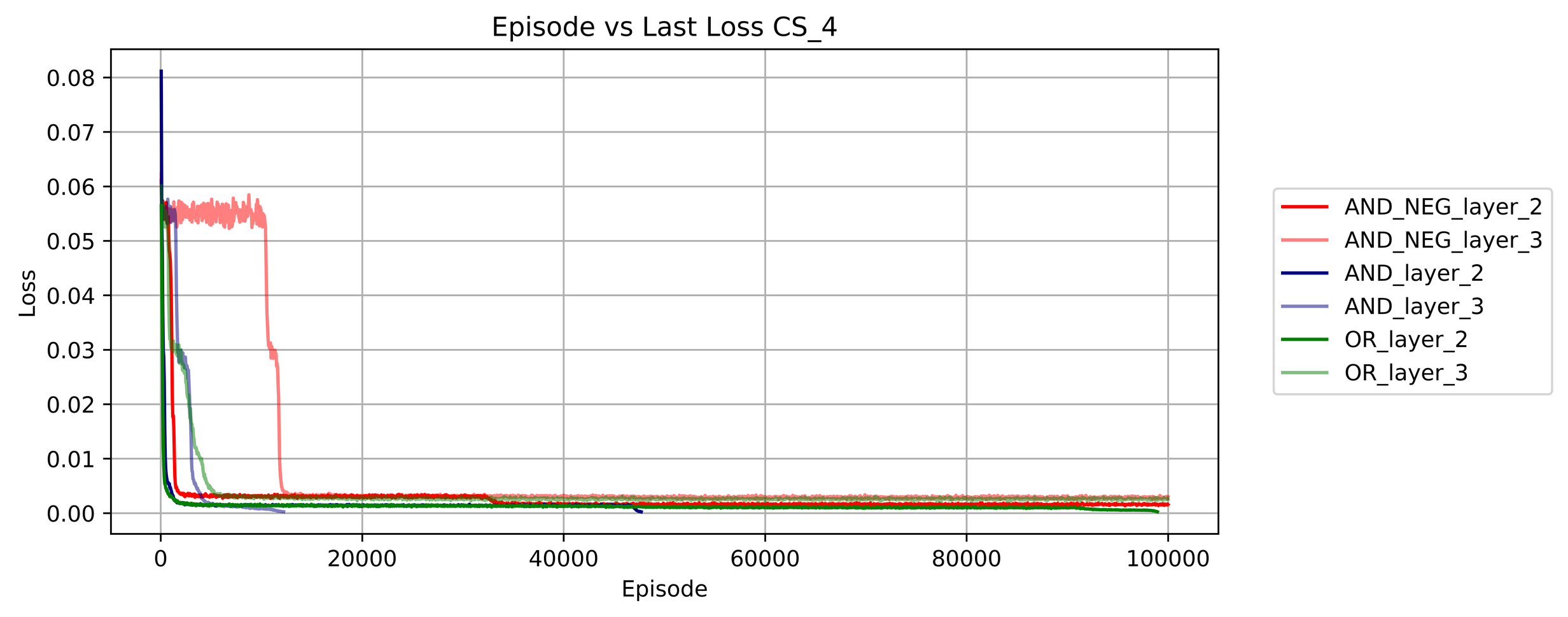}
    \caption{Comparison of loss functions for connection size 4.}
  \end{subfigure}
	
	\medskip

  \begin{subfigure}[t]{.48\textwidth}
    \centering
    \includegraphics[width=\linewidth]{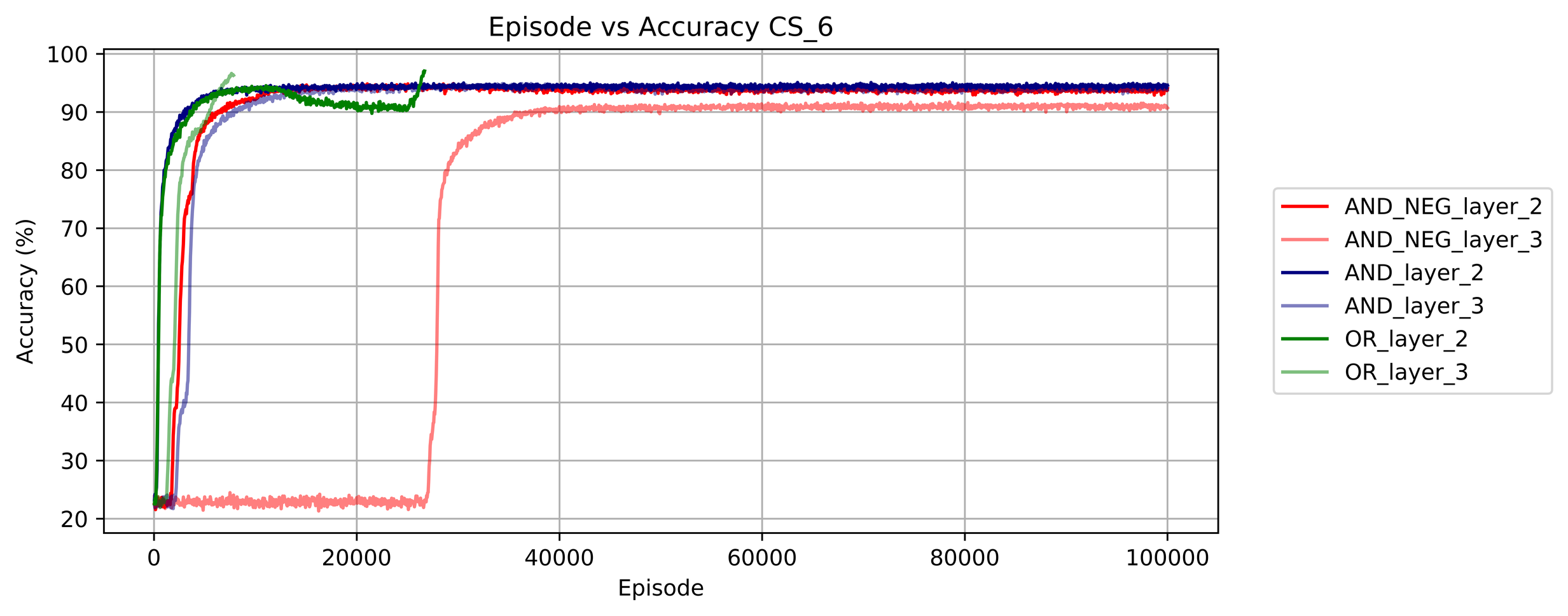}
    \caption{Comparison of testing accuracies for connection size 6.}
  \end{subfigure}
  \hfill
  \begin{subfigure}[t]{.48\textwidth}
    \centering
    \includegraphics[width=\linewidth]{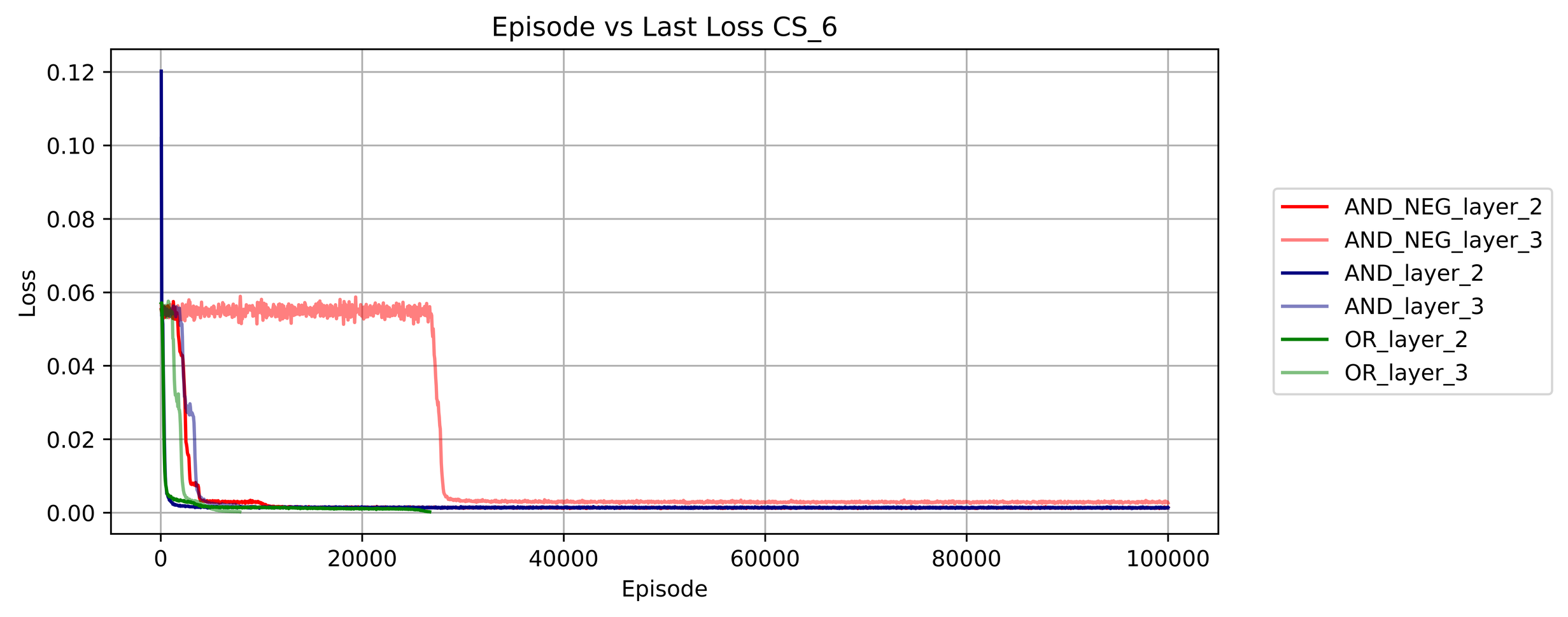}
    \caption{Comparison of loss functions for connection size 6.}
  \end{subfigure}
	
	\medskip

  \begin{subfigure}[t]{.48\textwidth}
    \centering
    \includegraphics[width=\linewidth]{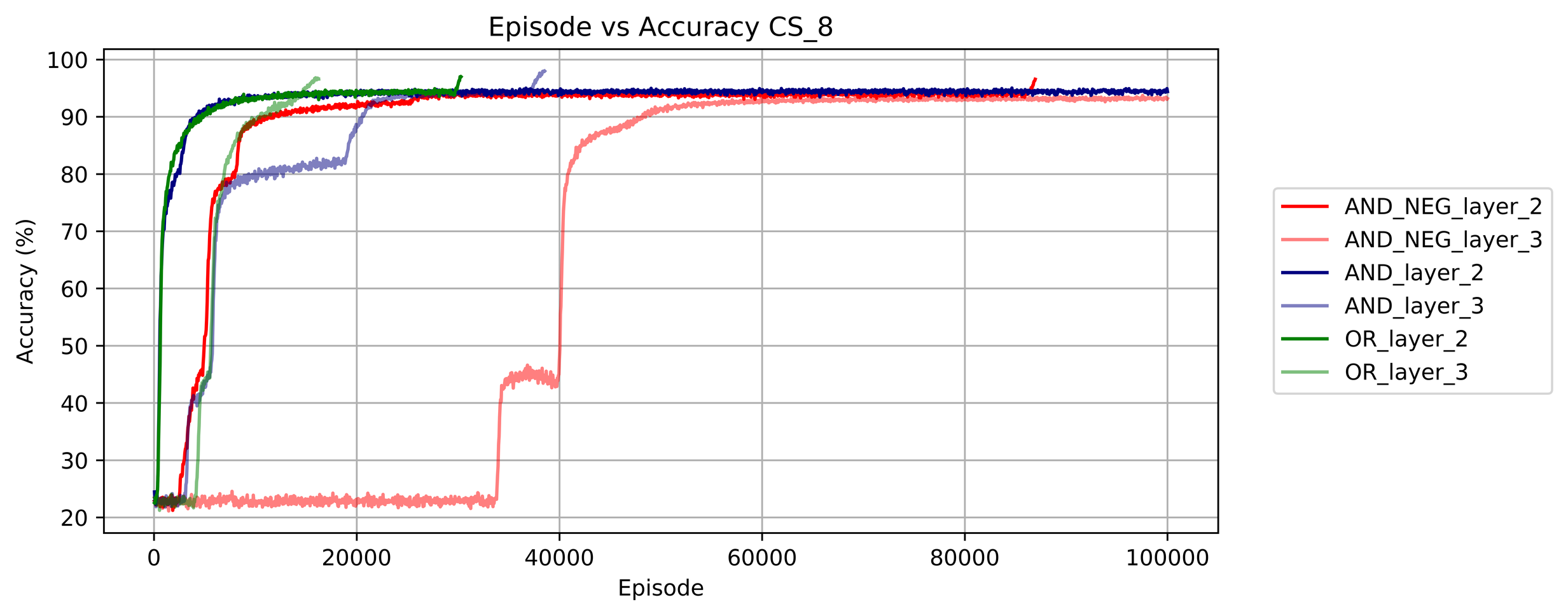}
    \caption{Comparison of testing accuracies for connection size 8.}
  \end{subfigure}
  \hfill
  \begin{subfigure}[t]{.48\textwidth}
    \centering
    \includegraphics[width=\linewidth]{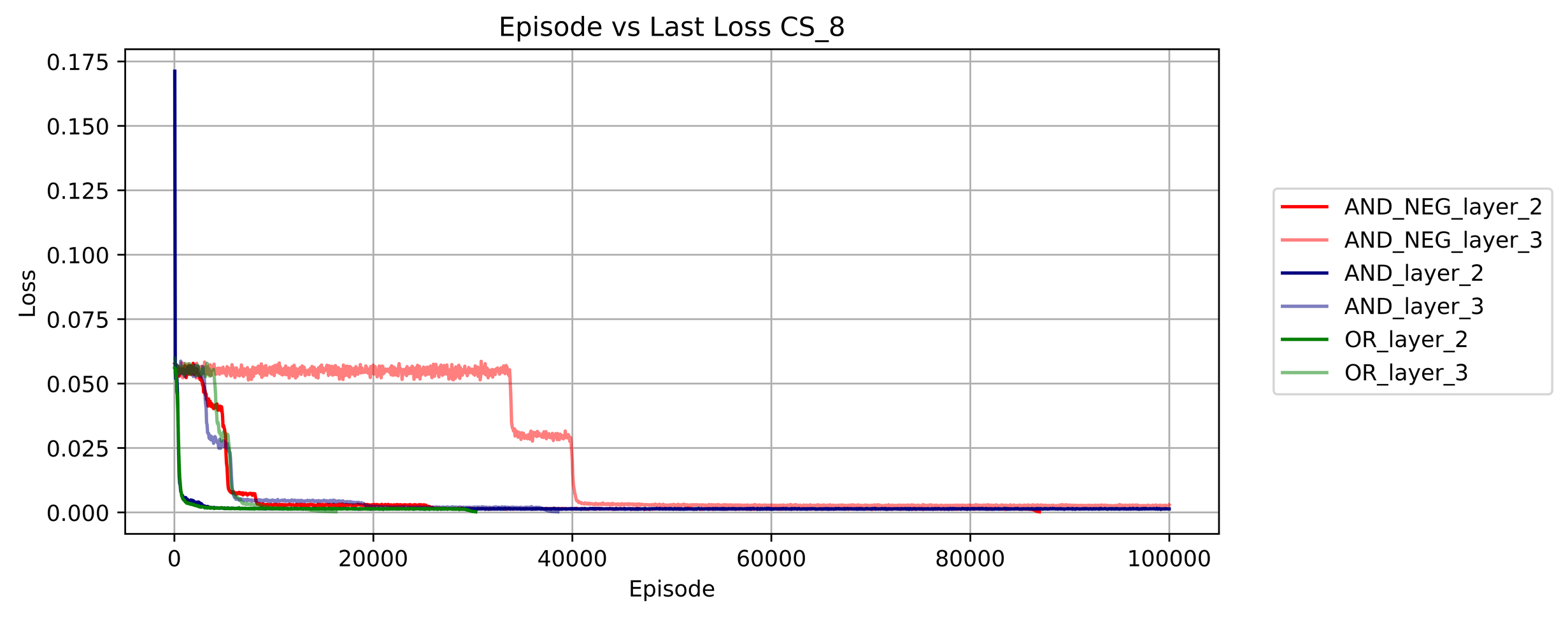}
    \caption{Comparison of loss functions for connection size 8.}
  \end{subfigure}
	
	\medskip

  \begin{subfigure}[t]{.48\textwidth}
    \centering
    \includegraphics[width=\linewidth]{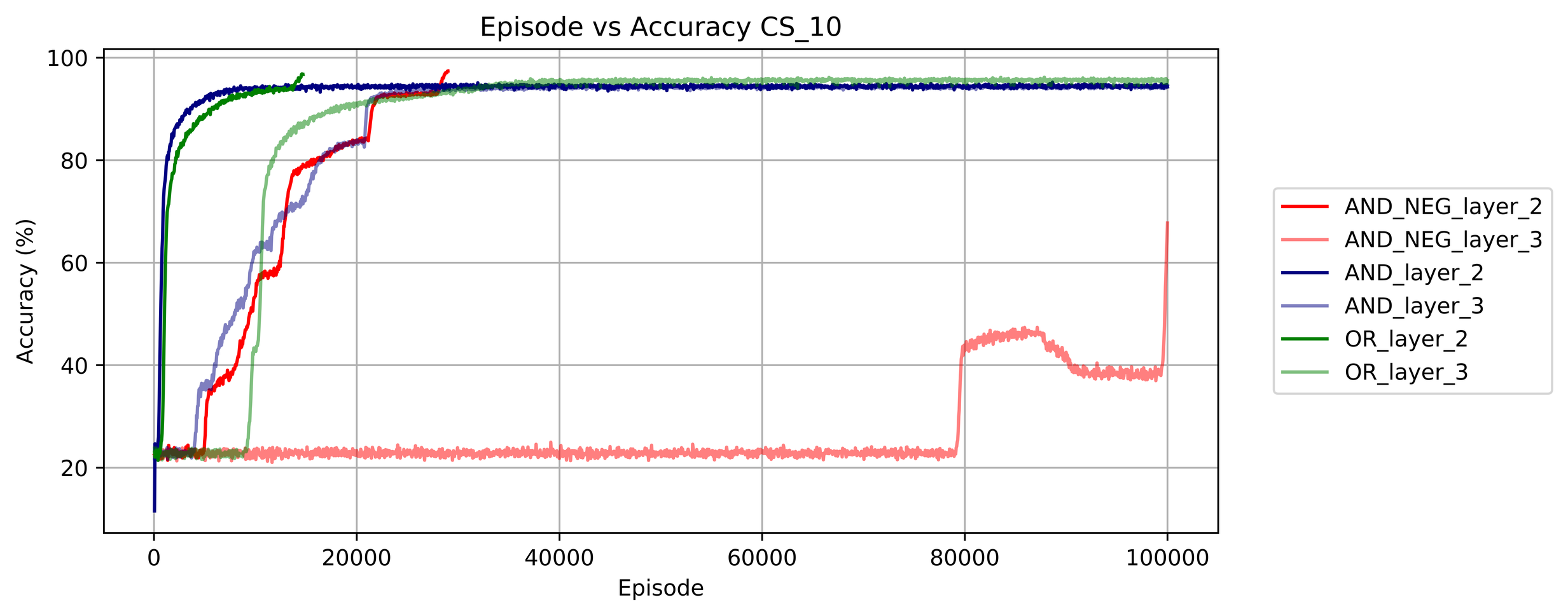}
    \caption{Comparison of testing accuracies for connection size 10.}
  \end{subfigure}
  \hfill
  \begin{subfigure}[t]{.48\textwidth}
    \centering
    \includegraphics[width=\linewidth]{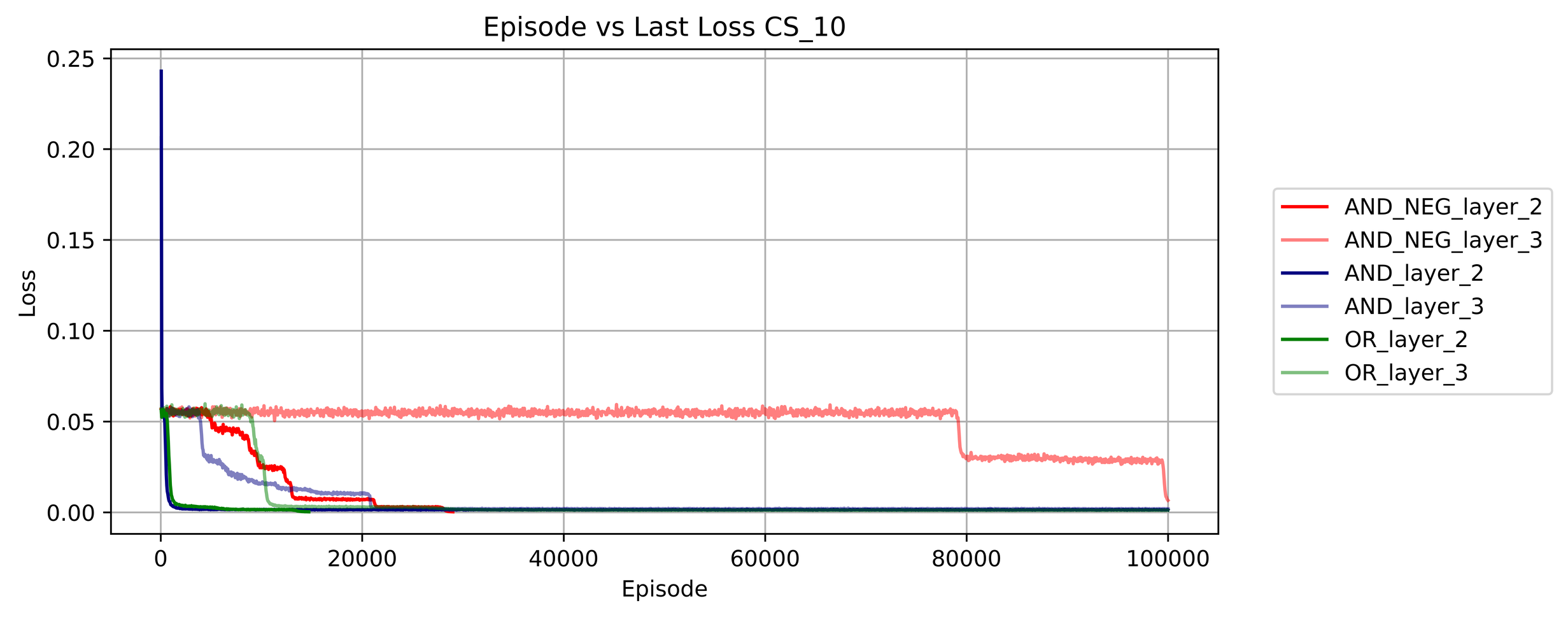}
    \caption{Comparison of loss functions for connection size 10.}
  \end{subfigure}
	\caption{Comparison of testing accuracy and loss function for different connection sizes and network architectures.}
	\label{fig:results}
\end{figure}

We compare three different NLRL structures on the data set, each with input layer of size two, two or three NLRL respectively, output layer of size 10 and varying numbers of neurons per layer. First, we employ NLRL with inbuilt AND and OR-relations (termed AND-OR) and corresponding output gating. Second, we employ NLRL in negation normal form, i.e. using AND-relations and negation at input and output only (termed AND-NEG). Third, we remove the negation gating of the output as this creates some redundancy of the negation operation when NLRL are stacked (termed AND-NONEG). In fact, the output negation and subsequent input negation can both change their weights without affecting the results. We test with different numbers of neurons, also termed connection size (CS), per logic rule layer, namely with CS of 2, 4, 6, 8 and 10. Hence the network size is 2-CS-(CS)-10. We employ a learning rate of 10 with a minibatch size of 20 and put $\epsilon=10^{-5}$ for numerical stability. Neither batch normalization nor dropout is used. The networks are initialized randomly by uniformly sampling from the interval $[-0.5, \ 0.5]$. The results of the network training are presented in Fig.~\ref{fig:results}. NOte, that we employ early-stopping which is why some curves suddenly stop before the maximum iteration.

As can be seen in the results, the AND-OR network architectures consistently converges earlier to optimal results than networks without the option to represent OR-relations. However this comes with the cost of higher computation times as shown later. This is obviously due to the better representation abilities of these networks. Thus, networks with only AND-operations (AND-NEG and AND-NONEG) have to model OR-relations by orchestrating negation and AND-relations accordingly which requires more training. Among the networks solely using the AND-operation, we can infer that networks without negation gating (AND-NONEG) perform better than with negation (AND-NEG). Particularly, the convergence speed is drastically reduced, especially with higher network sizes. As discussed these networks have inbuilt redundancy in the negation operation, if multiple layers are stacked. This makes training more difficult especially as a cyclic behavior during training can occur. 

In general, networks with three layers converge slower compared to 2-layer architectures which is not surprising due to the higher numbers of parameters. The final results are better compared to 2-layer architectures. By considering the networks size, which is represented mainly by the connection size (CS) between subsequent layers, we can see that the performance increases with increasing CS until a saturation occurs at CS 8. Increasing the CS further does not increase the performance indicating that the necessary network size has been reached.

Considering the course of the loss function over iterations, we observe a stepwise behavior. This is caused due to the combinatorical nature of the optimization problem such that a reduction of the loss is achieved if particular combination of boolean parameters converge to specific values. This is most likely amplified due to the derivative of the sigmoid function, enabling big parameter changes in a region around zero. Hence, the steps can be related to individual rules having converged.

\begin{figure}[h]
 \centering
 \includegraphics[width=\columnwidth,keepaspectratio]{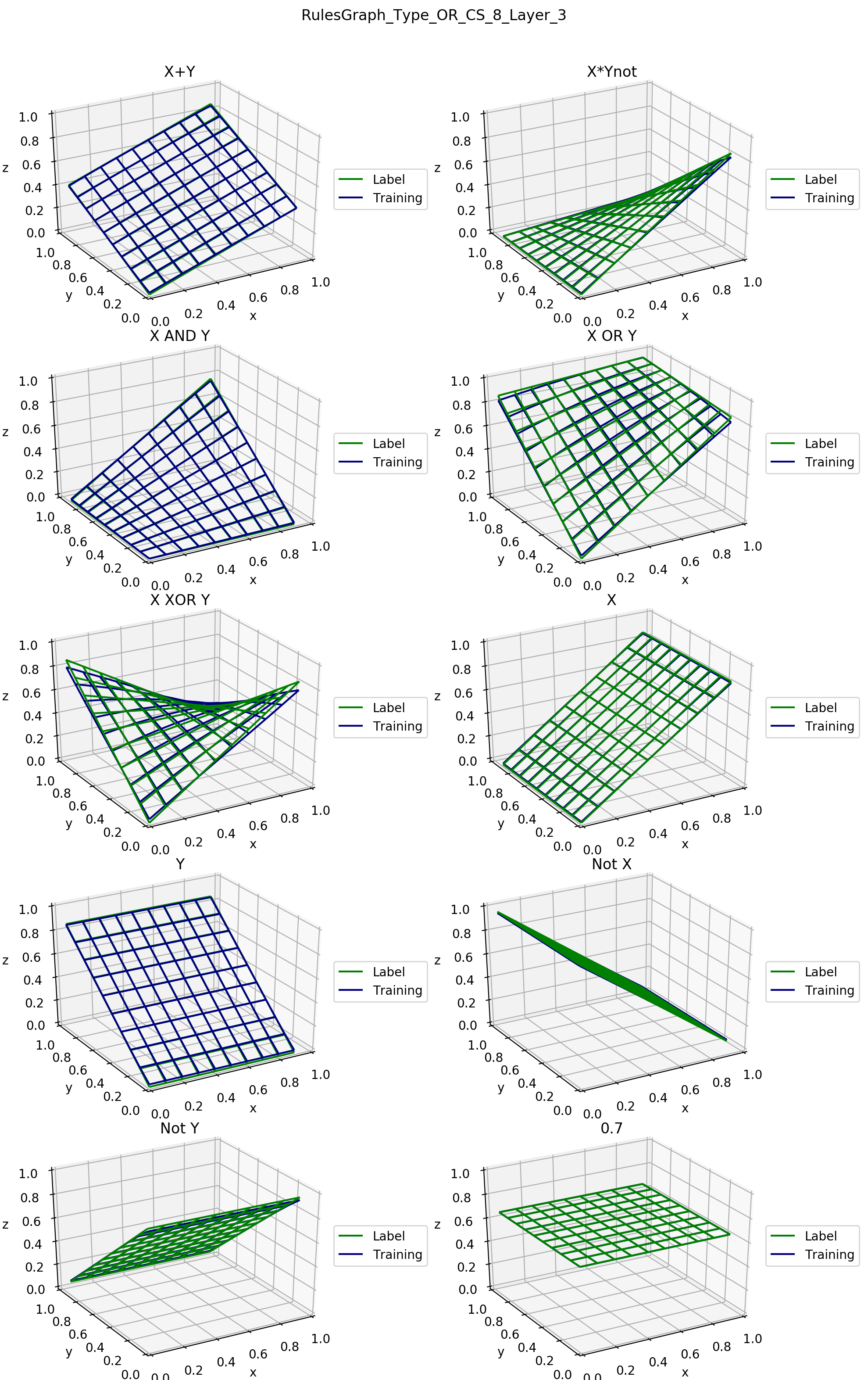}
\caption{Illustrations of the learned surfaces representing the logic and arithmetic relations for the network architecture 2-8-8-10.}
\label{fig:headmap}
\end{figure}
Fig.~\ref{fig:headmap} shows the learned surfaces of the logic and arithmetic relations to be trained for one of the best performing networks to give a qualititative representation of the learned manifold, which underlines the performance of the network on the test data set.
 
\begin{figure}[h]
 \centering
 \includegraphics[width=\columnwidth,keepaspectratio]{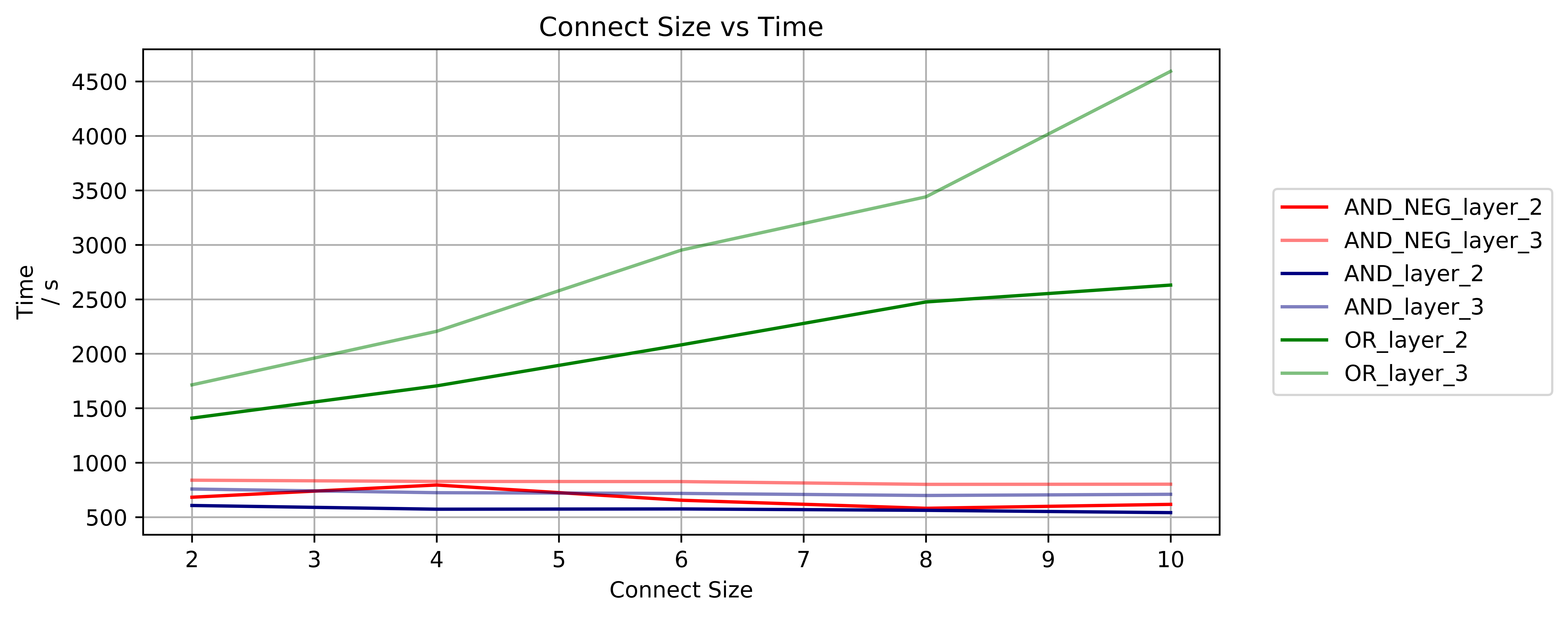}
\caption{Training times for the various architectures depending on the connection size.}
\label{fig:comptimes}
\end{figure}

Finally, we compare the different network architectures concerning their computation times in Fig.\ref{fig:comptimes}.
As already stated, the AND-OR net has a considerable higher computation times compared to the AND-NEG and AND-NONEG networks.


\section{Conclusion}

We presented neural logic rule layers which are able to represent arbitrary logical relations using propositional logic and negation gating. The proposed network layer is generally applicable and can be used in arbitrary network architectures. Furthermore, the proposed architecture allows for a direct incorporation of available human knowledge in form of predefined rules. NLRL have some nice properties, i.e. they are able to not only represent logical rules but also certain types of nonlinear functions and share some relations to fuzzy logic. In fact, depending on the chosen input activation function which can be considered as membership functions in the context of fuzzy logic, the architecture allows for an end-to-end training of the rules parameter, the logic relations and the number of active rules. The results applied on some synthetic data sets reveal the applicability of the proposed architecture. 

In future research we will apply and test NLRL on various data sets, including image classification. Furthermore, we will explore the incorporation of NLRL in recurrent neural network architectures. This would result in RNN structures representing finite state automata which is especially applicable for data analytics and reinforcement learning in discrete environments like discrete production systems.

\end{document}